\pdfoutput=1
\documentclass[11pt]{article}
\usepackage[final]{acl}
\usepackage{times}
\usepackage{latexsym}
\usepackage{array}
\usepackage{float} 
\usepackage{multirow}
\usepackage{natbib}
\usepackage{tabularx}
\usepackage{amssymb}
\usepackage{makecell}
\usepackage{pifont}
\usepackage{booktabs}
\usepackage{xcolor}
\usepackage{booktabs}
\usepackage{graphicx}
\definecolor{darkcheck}{RGB}{0,200,0}  
\definecolor{darkcross}{RGB}{200,0,0}  
\newcommand{\cmark}{\textcolor{darkcheck}{\ding{51}}}
\newcommand{\xmark}{\textcolor{darkcross}{\ding{55}}}

\usepackage[T1]{fontenc}
\usepackage[utf8]{inputenc}
\usepackage{microtype}
\usepackage{inconsolata}
\usepackage{bbding}
\usepackage{graphicx}
\usepackage[ruled,vlined,linesnumbered]{algorithm2e}
\usepackage{amsmath,amssymb}

\SetKwComment{Comment}{// }{} 
\SetKwInput{KwData}{Input}
\SetKwInput{KwResult}{Output}

\DontPrintSemicolon

\definecolor{aclgreen}{rgb}{0.15, 0.55, 0.15} 
\definecolor{aclred}{rgb}{0.7, 0.15, 0.15}  
\title{MERIT: Multilingual Expert-Reward Informed Tuning for Chinese-Centric Low-Resource Machine Translation}

\author{
  Zhixiang Lu, Chong Zhang, Chenyu Xue, \\
  \textbf{Angelos Stefanidis, Chong Li, Jionglong Su, Zhengyong Jiang\textsuperscript{\Envelope}} \\
  Xi’an Jiaotong-Liverpool University \\
  \texttt{zhengyong.jiang02@xjtlu.edu.cn}
}

\begin{document}

\maketitle
\begin{abstract}
Neural machine translation (NMT) from Chinese to low-resource Southeast Asian languages remains severely constrained by the extreme scarcity of clean parallel corpora and the pervasive noise in existing mined data. This chronic shortage not only impedes effective model training but also sustains a large performance gap with high-resource directions, leaving millions of speakers of languages such as Lao, Burmese, and Tagalog with persistently low-quality translation systems despite recent advances in large multilingual models. We introduce \textbf{M}ultilingual \textbf{E}xpert-\textbf{R}eward \textbf{I}nformed \textbf{T}uning (\textbf{MERIT}), a unified translation framework that transforms the traditional English-centric ALT benchmark into a Chinese-centric evaluation suite for five Southeast Asian low-resource languages (LRLs). Our framework combines language-specific token prefixing (LTP) with supervised fine-tuning (SFT) and a novel group relative policy optimization (GRPO) guided by the semantic alignment reward (SAR). These results confirm that, in LRL{\textrightarrow}Chinese translation, targeted data curation and reward-guided optimization dramatically outperform mere model scaling.
\end{abstract}

\section{Introduction}
\label{sec:intro}

The vision of Neural Machine Translation (NMT) is to provide equitable access to information for speakers of over 7,000 languages worldwide. However, the benefits brought by recent advances in large-scale models are highly uneven. High-resource language pairs such as English–French have achieved near-human BLEU scores \citep{papineni2002bleu}, while many low-resource languages remain almost entirely untranslatable due to severe data scarcity \citep{nllb2024scaling}.

This disparity is marked between Chinese and low-resource languages (LRLs), spanning domestic (e.g., Burmese, Lao). These languages suffer from scarce parallel and monolingual data, limited annotators, and sometimes non-standardized orthographies, deepening the digital divide. While multilingual models like mBART-50 \citep{liu2020mbart} and mT5 \citep{xue2021mt5} show zero-shot gains, they still underperform on these LRLs. Even with NLLB-200's expanded coverage \citep{nllb2024scaling}, LRL-Chinese performance continues to trail English-pivoted directions.

Moreover, the lack of publicly available and high-quality evaluation benchmarks hinders objective progress measurement. An ideal benchmark should: (i) cover multiple LRL{\textrightarrow}Chinese directions directly to avoid pivot bias, (ii) maintain sufficient balance in scale and domain coverage, (iii) avoid dependence on English as a pivot. Without these features, improvements in Chinese-centric translation are difficult to reproduce or attribute reliably. To address these challenges, this paper makes the following three contributions:

\begin{itemize}
  \item \textbf{Benchmark Construction:} We introduce \textbf{CALT}, the first Chinese-centric benchmark for Southeast Asian languages. By reconstructing the ALT corpus to eliminate English-pivot bias, we establish a rigorous, pivot-free standard for evaluating direct LRL-Chinese translation.

  \item \textbf{Methodological Innovation:} We propose \textbf{MERIT}, a framework that combines LTP, SFT, and a novel GRPO method with SAR for efficient data filtering and model refinement.
  \item \textbf{Empirical Validation:} Experiments show that MERIT-3B, using only 22.8\% of the original data, significantly outperforms much larger baselines, showing that high-quality data and reward-guided optimization trump model scale in low-resource translation.
\end{itemize}

\section{Related Work}
\paragraph{LRL{\textrightarrow}Chinese Corpora.} The CCMT shared tasks released fewer than 200k sentence pairs for \textsc{zh--ug} and \textsc{zh--mn} \citep{liu2021ccmt}, while Wiki-based mining typically yields only a few thousand pairs for \textsc{zh--lo} and \textsc{zh--fil} \citep{artetxe2019massively}. The ALT corpus \citep{thu2016alt} extends coverage to 13 LRLs but remains English-centric and lacks direct LRL{\textrightarrow}Chinese alignment.

\paragraph{LLM-based Machine Translation.} In recent years, the integration of LLMs into machine translation has witnessed rapid and substantial progress. Related research mainly focuses on low-resource scenarios \cite{zebaze-etal-2025-compositional,lu2025advancing}, context and prompt optimization \cite{feng-etal-2024-ladder,zaranis-etal-2024-analyzing}, self-refining mechanism \cite{feng-etal-2025-tear,wang-etal-2024-mitigating-language}, term control \cite{kim-etal-2024-efficient}, dialect and historical language translation \cite{abdelaziz-etal-2024-llm,volk-etal-2024-llm}, document-level translation \cite{wang-etal-2023-document-level}, simultaneous translation \cite{koshkin-etal-2024-transllama}, and fusion of traditional MT methods \cite{hoang-etal-2024-fly}. These works demonstrate that LLM has significantly outperformed traditional methods in various challenging scenarios through a combination of translation, structured reasoning, self-refinement, and targeted fine-tuning techniques, while providing efficient and feasible solutions for low-resource languages, specialized domains, and real-time translation.

\paragraph{Multilingual Pretraining Models.} Multilingual pretrained models aim to learn shared representations across tens to hundreds of languages with a single LLM, enabling cross-lingual transfer with zero-shot/few-shot tasks\cite{lauscher-etal-2020-zero}. Early representative works include mBART-50 \citep{liu2020mbart}, mT5 \citep{xue2021mt5} and DeltaLM \citep{ma2021deltalm}, which cover 50 to 101 languages, respectively. NLLB-200 \citep{nllb2024scaling}, released by Meta in 2022, achieves high-quality mutual translation of 200 languages for the first time. Through technologies such as massively parallel data mining, back-translation, and self-supervised denoising, the average BLEU score on the FLORES-200 benchmark has been improved by 44\% compared to the best system at the time \cite{nllb2024scaling}. Additionally, the XSTS manual evaluation index is introduced, which is more closely aligned with human judgment \cite{nllb2024scaling}. However, our fine-grained analysis reveals that a significant quality gap still exists in the LRL-to-Chinese direction, underscoring the importance of parallel data quality, domain distribution, and the cross-lingual alignment strategy, as much as model size itself.

\section{Methodology}

\begin{figure*}[htbp]
 \vspace{-20pt}
 \includegraphics[width=1.0\linewidth]{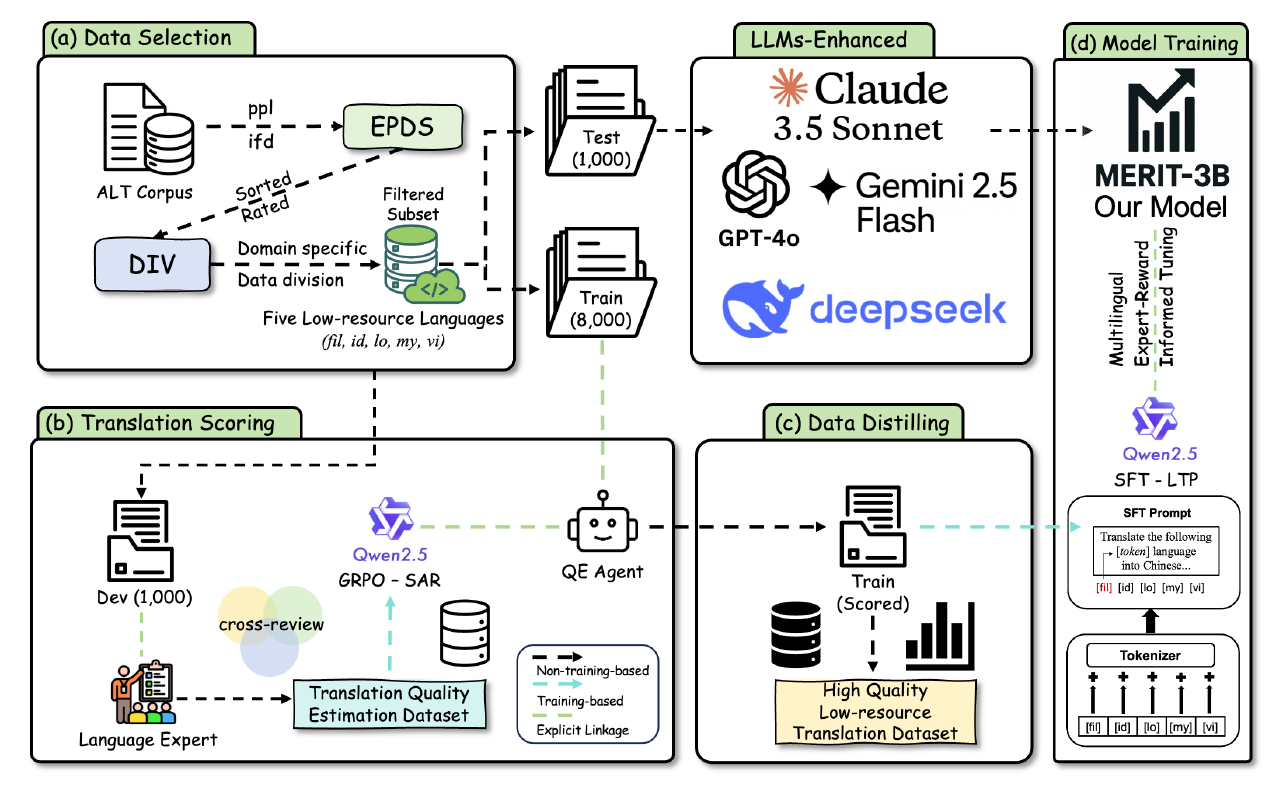}
 \vspace{-20pt}
\caption{Overview of the MERIT framework. The pipeline consists of (a) heuristic data selection utilizing the Elite Parallel Data Sampler (EPDS, \autoref{alg:epds}) and Data Integrity Validation (DIV, \autoref{alg:div}), (b) translation scoring via a QE agent trained on expert cross-reviewed quality data using GRPO-SAR, (c) reward-based data distillation, which utilizes the QE agent to rescore the entire training set and filter for high-quality translation data, and (d) final model optimization using SFT-LTP based on this distilled dataset.}
 \label{fig:framework}
 \vspace{-15pt}
\end{figure*}
\subsection{Supervised Fine-Tuning}
We fine-tune open-source models (Qwen-2.5-0.5B and 3B) on the CALT benchmark described in \autoref{sec:calt_bench} using SFT \citep{fan2021beyond}. The objective of SFT is to maximize the conditional probability of producing the target sequence $Y=(y_1, \dots, y_M)$ given the source sequence $X=(x_1, \dots, x_N)$, following the standard sequence-to-sequence formulation \citep{sutskever2014seq2seq}. 
During training, we employ a teacher-forcing strategy \citep{williams1989learning, lu2026sage}, conditioning the model on the ground-truth previous tokens $y_{<t}$ to predict the next token $y_t$. The model is trained by minimizing the cross-entropy loss. To prevent over-confidence and improve generalization, we incorporate label smoothing \citep{szegedy2016rethinking, lu2026}. The token-level loss is defined as:
\begin{equation}
\label{eq:sft_split_aligned_number}
\small
\begin{aligned}
  \mathcal{L}_{\text{SFT}}(y_t, \hat{P}_t) &= - \sum_{k=1}^{|V|} q'(k|y_t^*) \log P(k|X,y_{<t};\theta) \\
  q'(k|y_t^*) &= (1-\varepsilon_{\text{ls}})\mathbb{I}(k=y_t^*) + \frac{\varepsilon_{\text{ls}}}{|V|}
\end{aligned}
\end{equation}
Fine-tuning is performed separately for each LRL-Chinese pair to respect language-specific morphology. This combination of filtering and task-specific adaptation yields substantial gains over zero-shot baselines, echoing findings on targeted adaptation for massively multilingual models \citep{arivazhagan2019massively}.

\subsection{Language-specific Token Prefixing}
We introduce a Language-specific Token Prefixing (LTP) strategy to improve language discrimination in many-to-one multilingual generation. This method involves injecting a language-identifier token into both the tokenizer vocabulary and the SFT prompt, enabling consistent language conditioning. Given a sample consisting of a source sequence $X$ (from LRLs) and a target sequence $Y$ (Chinese). We prepend the specific language token corresponding to the source, denoted as $[lang]$, to the source input $X$ to create a modified source $X^{\prime}$ as shown in Equation~\ref{eq:LTP_x_prime_pdf}:
\begin{equation}
  \small
  X^{\prime} = [lang] \oplus X = [lang, x_{1}, x_{2}, ..., x_{n}] \\ \label{eq:LTP_x_prime_pdf}
\end{equation}
A prompt instruction $P(l)$ is defined for the target language $l$. An example structure for such a prompt is:
\begin{equation}
  \small
  P(l) = \text{''Translate [\texttt{lang}] language into Chinese: ...''} \label{eq:LTP_prompt_pdf}
\end{equation}
The final input to the model is then constructed by combining the prompt and the original source input. Based on the structure presented, the final model input is:
\begin{equation}
  \small
  \text{Input} = [p_{1},...,p_{r},\textit{lang},x_{1},...,x_{n}] \label{eq:LTP_final_input_pdf}
\end{equation}
where $[p_{1},...,p_{r}]$ are the tokens derived from the prompt $P(l)$, and $[l]$ is the language identifier token explicitly prepended before the source tokens $X$.

The training objective is to minimize the negative log-likelihood of the target sequence $Y$, conditioned on the prompt $P(l)$ and the source input $X$:
\begin{equation}
  \small
  \mathcal{L}_{\text{MLE}} = -\sum_{t=1}^{m} \log P(y_{t} | y_{<t}, P(l), X; \theta) \label{eq:LTP_mle_loss_pdf}
\end{equation}
where $y_{<t}$ represents the preceding ground-truth target tokens, and $\theta$ denotes the model parameters.

This LTP method extends the target language prefixing idea introduced by \citep{johnson2017googlenmt} and adapts the prompt-based control framework from \citet{Tan2023MultilingualTV}. By combining tokenizer-level symbolic conditioning with prompt-level natural language alignment, tailored for unified training in a one-to-many multilingual setup.

\subsection{Group Relative Policy Optimization}
Group Relative Policy Optimization (GRPO) is a reinforcement learning strategy that refines model outputs using reward feedback. Inspired by Reinforcement Learning with Human Feedback techniques \citep{ouyang2022training}, GRPO operates on mini-batches of candidate translations in this task, assigning scalar rewards based on SAR scores. Subsequently, the model learns to maximize the expected reward via policy gradient updates. Unlike conventional pointwise objective functions, GRPO introduces an intra-batch comparison mechanism and normalizes rewards using a moving baseline. This approach helps to reduce the variance of gradient estimates, thereby enhancing the stability of the training process \cite{lu2026sage}. Our experiments indicate that GRPO is effective for translation evaluation, enabling the selection and improvement of translation quality in datasets.

\subsection{Semantic Alignment Reward Function}
\label{sec:sar_function}

We introduce the Semantic Alignment Reward (SAR) to align the Quality Estimation (QE) agent's predictions with human expert judgments. SAR incentivizes the agent to generate scalar quality scores that precisely reproduce ground-truth expert values, ensuring the reliability of the reward signal used in downstream policy optimization.

Given the QE agent's generated evaluation log $c_i$, we extract potential quality scores using a predefined regular expression $R$ (e.g., capturing patterns like ``Score: [0-100]''). Let $M(c_i)$ denote the set of all integer matches found within $c_i$:
\begin{equation}
\label{eq:match_set_M_ci}
  \small
  M(c_i) = \{m \in \mathbb{Z} \mid P_R(m, c_i)\}
\end{equation}
where $P_R(m, c_i)$ is true if integer $m$ matches $R$ in $c_i$. To resolve ambiguity in generated logs, we employ a conservative extraction function $E(c_i)$ that selects the minimum valid integer found, thereby penalizing outputs containing conflicting lower scores. If no match exists, a penalty indicator of -1 is assigned:
\begin{equation}
\label{eq:sar_extractor_si}
  \small
  s_i = E(c_i) =
  \begin{cases}
    \min M(c_i), & \text{if } M(c_i) \neq \emptyset \\
    -1,      & \text{if } M(c_i) = \emptyset
  \end{cases}
\end{equation}

Let $a_i$ represent the ground-truth expert score. We define the reward based on the absolute deviation $d = |s_i - a_i|$. To allow for minor subjective variations in human annotation while enforcing strict accuracy, we employ a stepwise mapping $\phi(d)$:
\begin{equation}
\label{eq:sar_phi_mapping}
\small
\phi(d) =
  \begin{cases}
    2.0, & \text{if } d = 0 \\
    1.0, & \text{if } 1 \le d \le 10 \\
    0.0, & \text{otherwise}
  \end{cases}
\end{equation}
This mapping grants maximum reward for exact matches and partial reward for acceptable estimations (within a 10-point tolerance), while zeroing out significant misjudgments.

The final alignment reward $r_i$ is computed by applying this mapping to valid extractions. Instances where the agent fails to generate a parseable score ($s_i < 0$) receive zero reward:
\begin{equation}
\label{eq:sar_reward_ri}
  \small
  r_i =
  \begin{cases}
    \phi(|s_i - a_i|), & \text{if } s_i \ge 0 \\
    0.0,       & \text{if } s_i < 0
  \end{cases}
\end{equation}
This mechanism provides a dense supervision signal, effectively steering the QE agent to approximate the distribution of human expert preferences \citep{wu2023finegrained}.



\subsection{Dataset}
\label{sec:calt_bench}
We construct a new test suite based on the ASEAN Languages Treebank (ALT) corpus \citep{thu2016alt}, called CALT. ALT is an English-centric multilingual corpus that already provides sentence-level alignment for several Southeast-Asian languages-Vietnamese (\texttt{vi}), Burmese (\texttt{my}), Lao (\texttt{lo}), Tagalog (\texttt{fil}), and Indonesian (\texttt{id}), the details shown in \autoref{tab:alt-stats}. Although Chinese is included as a target aligned with English, no direct LRL{\textrightarrow}Chinese alignment exists. We therefore re-index sentences that share the same \texttt{alt\_id} and semantic source to form direct LRL-Chinese sentence pairs. In the resulting test set, Chinese can serve as either the source or the reference language.
\begin{table}[ht]
\vspace{-1em}
\centering
\small
\resizebox{0.48\textwidth}{!}{
\begin{tabular}{lcccc}
\toprule
\textbf{Languages} &
\makecell{\textbf{Speaker Population}\footnotemark[1]\\~\citep{ethnologue2023}} &
\makecell{\textbf{Filtered}\\ \textbf{Subset}\footnotemark[2]} & \textbf{LRL} \\
\midrule
Chinese  (\texttt{zh}) & 1180M & \ding{55}  & \xmark \\
Hindi   (\texttt{hi}) & 345M & \ding{55}  & \xmark \\
Bengali  (\texttt{bn}) & 234M & \ding{55}  & \xmark \\
Japanese  (\texttt{ja}) & 121M & \ding{55}  & \xmark \\
Vietnamese (\texttt{vi}) &  86M & 10K & \cmark \\
Indonesian (\texttt{id}) &  43M & 10K & \cmark \\
Burmese  (\texttt{my}) &  33M & 10K & \cmark \\
Tagalog  (\texttt{fil}) &  24M & 10K & \cmark \\
Thai    (\texttt{th}) &  20M & \ding{55}  & \xmark \\
Malay   (\texttt{ms}) &  18M & \ding{55}  & \xmark \\
Khmer   (\texttt{km})\footnotemark[3] &  16M & --  & \cmark \\
Lao    (\texttt{lo}) & 4.3M & 10K & \cmark \\
\bottomrule
\end{tabular}
}
\caption{The ALT corpus statistics sorted by L1 speaker population. All counts refer to L1 speakers and are rounded to the nearest million (M). }
\label{tab:alt-stats}
\vspace{-2em}
\end{table}

\subsection{Language Selection}

We deliberately focus on five Southeast Asian low-resource languages (Vietnamese, Burmese, Lao, Tagalog, and Indonesian) for four data-driven reasons. All five appear in ALT with reliable English alignments, making high-quality LRL-Chinese re-alignment feasible. We retain Indonesian instead of Malay because the two belong to the same Malayic subgroup with ~90\% lexical overlap and are often treated as a single “Malay macrolanguage” in international surveys \citep{adelaar2012lexical}, including both would introduce redundancy. Malay already has over 1 million clean En–Ms sentence pairs (e.g., MT-Wiki and multiple OPUS sub-corpora), placing it in the mid-resource tier \citep{duong2017mtwiki}, whereas Indonesian still lacks sizable Chinese parallel data (<50k pairs in total, with ALT contributing only 20k) and remains low-resource according to FLORES-200 and NLLB benchmarks \citep{goyal2022flores}. Thai has aggregated corpora exceeding one million pairs and dedicated WMT/IWSLT tracks \citep{lowphansirikul2020scb}, so it no longer meets a strict LRL definition. Khmer parallel resources are both small and highly noisy, with the WMT20 corpus-filtering task highlighting the need for extensive cleaning \citep{koehn2020wmt}. Moreover, Ethnologue reports virtually no L2 speaker community for Khmer \citep{ethnologue2023}, which would require a language-specific filtering pipeline and undermine comparability. This reconstructed benchmark complements existing resources such as FLORES-200 \citep{nllb2024scaling}, particularly for LRL-Chinese directions in mainland and maritime Southeast Asia. Unlike pivot-based benchmarks, our test set avoids semantic distortion introduced by intermediate English, thereby enabling more realistic, stable, and reproducible evaluation of Chinese-centric multilingual translation systems.

\footnotetext[1]{Speaker numbers derive from the most recent national censuses or \textit{Ethnologue} reports (2023–2025) and are expressed in millions (M).}
\footnotetext[2]{Each ALT language contains approximately 20k aligned sentence pairs from a shared English source. See \url{https://www2.nict.go.jp/astrec-att/member/mutiyama/ALT/}.}
\footnotetext[3]{No L2 speaker community \citep{ethnologue2023}.}

\subsection{Model Overview}
We evaluate a series of representative LLMs, spanning both proprietary and open-source systems:
\vspace{-5pt}
\begin{itemize}
  \item \textbf{Qwen-2.5} \citep{ghosal2024promptrefine, cui2025multilingual}: Chinese-English bilingual models fine-tuned for multilingual transfer, evaluated on several LRLs.
  \item \textbf{GPT-4o} \citep{huang2025benchmax}: OpenAI's flagship model tested on 16 languages, including several low-resource directions such as En--Te and En--Sw.
  \item \textbf{Claude-3.5} \citep{enis2024claude}: A multilingual LLM from Anthropic, evaluated via MQM metrics on pairs like En--Yoruba and En--Amharic.
  \item \textbf{Gemini-2.5} \citep{gemini2025blog}: While lacking peer-reviewed benchmarks on LRL{\textrightarrow}Chinese tasks, its predecessor covers ultra-low-resource translation (e.g., En--Kalamang).
  \item \textbf{DeepSeek} \citep{huang2025benchmax,jiang2025deepseek}: A competitive open-source model evaluated in the BenchMAX suite alongside GPT-4o.
\end{itemize}
Zero-shot prompting was applied exclusively to representative closed-source LLMs. For the Qwen-2.5 models, both SFT and GRPO-enhanced SFT were applied. Additional tests, as illustrated in the Data Distilling module of \autoref{fig:framework}, were conducted with SFT and GRPO-enhanced SFT using enhanced data. This data was derived from closed-source model outputs, filtered by the QE agent, which trained on expert-rated development sets and optimized with SAR to select only high-quality translations. This comparative design allows us to isolate the specific contributions of reward-guided data distillation, verifying whether high-quality data can effectively compensate for limited model scale. Model performance is assessed using standard metrics: BLEU-4 \citep{papineni2002bleu}, sacreBLEU \citep{post2018call}, chrF \citep{popovic2015chrf}, ROUGE-L \citep{lin2004rouge}, METEOR \citep{banerjee2005meteor}, and BERTScore \citep{zhang2019bertscore}.

\subsection{Scoring and Selection}

To construct high-quality parallel corpora for low-resource translation, we design a three-stage scoring and filtering pipeline that integrates interpretable statistical features with semantic evaluation, followed by reference-free quality estimation and threshold-based selection.

\textbf{Stage I: Statistical and Semantic Feature-Based Scoring.}
In the first stage, we perform multi-dimensional scoring to capture both surface-level consistency and deeper semantic alignment between source and target sentences. Specifically, we extract five statistical features: sentence length ratio, token count ratio, punctuation frequency difference, digit proportion difference, and lexical diversity difference. These features have proven effective in prior work on noisy parallel data detection and alignment assessment \cite{Munteanu2005, Sanchez-Cartagena2018}, particularly for identifying mismatches caused by language divergence or formatting inconsistencies.

To further enhance robustness, we incorporate two semantic signals. The first is conditional perplexity, which measures the fluency and naturalness of each sentence and is widely used for data selection in machine translation \cite{Moore2010, Junczys-Dowmunt2018}. The second is instruction-following discrepancy, which evaluates whether the target sentence faithfully follows the communicative intent of the source, inspired by recent advances in instruction tuning \cite{Li2023FromQuantityToQuality}. All features are normalized and integrated through a weighted combination, enabling the identification of superficially aligned but semantically mismatched sentence pairs \cite{Espla-Gomis2020}. Implementation details and ablation analysis are provided in the Appendix~\ref{sec:filtering_method}.

\textbf{Stage II: Reference-Free Quality Estimation.}
In the second stage, we adopt a quality estimation (QE) agent trained on expert-annotated development sets. Inspired by COMET-QE, this model provides reference-free assessments of semantic adequacy, fluency, and alignment quality. The QE agent refines the results of the initial feature-based scoring by evaluating translation candidates from a semantic perspective without requiring ground-truth references \cite{rei2020comet, freitag2021experts}.

\textbf{Stage III: Threshold-Based Distilling.}
In the final stage, we apply a filtering threshold to the QE scores. The threshold is empirically calibrated to balance high recall and quality retention, based on agreement with human evaluation on a validation set. Only sentence pairs exceeding the threshold are retained. This process ensures that the final training corpus is not only large in scale but also high in quality, making it suitable for fine-tuning compact models in low-resource scenarios.

\section{Experiments and Analysis}

\subsection{Evaluation Method}

We evaluate translation performance using both overlap-based and semantic-aware metrics:

\textbf{Strict-Overlap:} \textbf{BLEU-4} \citep{papineni2002bleu}, \textbf{sacreBLEU} \citep{post2018call}, and \textbf{ROUGE-L} \citep{lin2004rouge} assess lexical match and n-gram precision, which are crucial for evaluating surface-level accuracy and fluency.

\textbf{Semantic-Friendly:} \textbf{chrF} \citep{popovic2015chrf}, \textbf{METEOR} \citep{banerjee2005meteor}, and \textbf{BERTScore} \citep{zhang2019bertscore} measure semantic similarity and fluency robustness, capturing aspects that n-gram overlap alone might miss.

Each metric is computed on the reconstructed ALT test suite for five LRL{\textrightarrow}Chinese pairs. We report averages across directions, comparing zero-shot prompting, SFT, and GRPO-enhanced regimes.

To provide a balanced evaluation that captures both lexical precision and semantic adequacy, we propose a composite metric, \textbf{BLEU-chrF}. This metric integrates insights from both the Strict-Overlap and Semantic-Friendly categories of evaluation measures by taking the arithmetic mean of the BLEU-4 score and the chrF score:
\begin{equation}
\label{eq:bleu_chrf}
  \small
  \text{BLEU-chrF} = \frac{\text{BLEU-4} + \text{chrF}}{2}
\end{equation}

By averaging these two widely-used metrics, one emphasizing n-gram precision and the other character n-gram recall and F-score. We aim to achieve a more holistic assessment of translation quality, particularly for tasks where both lexical fidelity and semantic resemblance are important (see \autoref{tab:metric-comparison}).

\subsection{Main Result}
\begin{figure}[t]
 \includegraphics[width=\columnwidth]{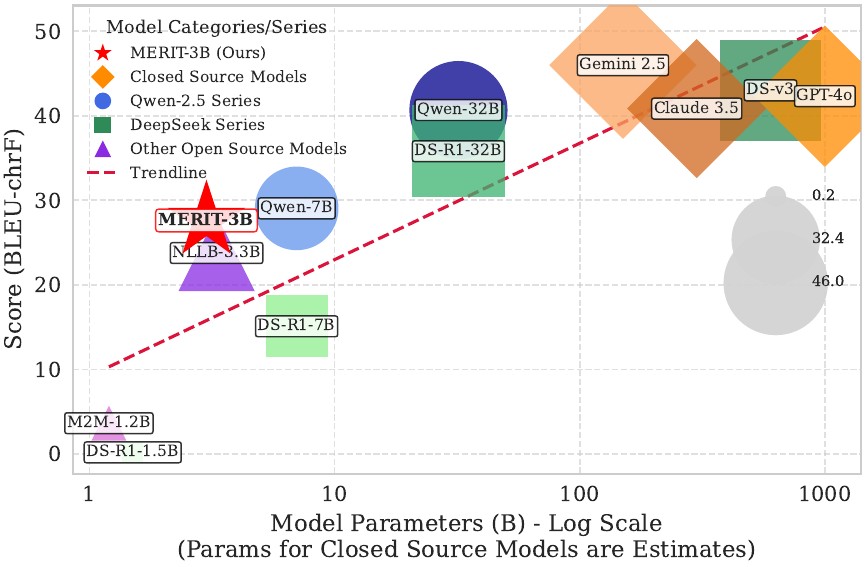}
 \vspace{-15pt}
 \caption{Performance–Scale Trade-offs of MERIT-3B and Baseline Models on Chinese-Centric Multilingual Translation. Comparison of BLEU-chrF scores against model size (log-scale) across MERIT-3B, open-source, and estimated closed-source models.}
  \label{fig:experiments}
  \vspace{-10pt}
\end{figure}
\begin{table*}[t]
\centering
\scriptsize
\setlength{\tabcolsep}{4pt}
\vspace{-12pt}
\resizebox{\textwidth}{!}{
\begin{tabular}{l|ccccc|ccccc|ccccc|c|c}
\toprule
\multicolumn{1}{c|}{\textbf{Strict-Overlap}} 
& \multicolumn{5}{c|}{\textbf{BLEU-4}} 
& \multicolumn{5}{c|}{\textbf{sacreBLEU}} 
& \multicolumn{5}{c|}{\textbf{ROUGE-L}} 
& \multicolumn{1}{c}{\textbf{}}
& \multicolumn{1}{c}{\textbf{}}\\
\cmidrule(r){1-1} \cmidrule(lr){2-16} \cmidrule(lr){17-18}
\multicolumn{1}{c|}{Model} 
& fil & id & lo & my & vi
& fil & id & lo & my & vi
& fil & id & lo & my & vi
& FT 
& OS \\
\midrule
GPT-4o     & 45.26 & 47.77 & 28.10 & 30.48 & 45.03     & 43.60 & 45.81 & 27.38 & 29.53 & 44.50 & 33.62 & \underline{31.40} & 12.33 & \underline{23.57} & 28.25     & \xmark & \xmark \\
Claude-3.5 Sonnet & 42.97 & 47.35 & \underline{32.20} & \underline{30.92} & \underline{45.14}  & 43.38 & 46.54 & \underline{32.65} & \underline{30.14} & \underline{44.55} & 35.03 & 31.17 & \textbf{14.00} & \textbf{24.44} & 28.45     & \xmark & \xmark \\
Gemini-2.5 Flash& \textbf{49.26} & \textbf{48.73} & \textbf{35.79} & \textbf{36.91} & 39.24 & \textbf{47.48} & \textbf{47.20} & \textbf{35.03} & \textbf{36.07} & 38.63 & \textbf{35.89} & 30.83 & \underline{13.53} & 23.92 & 24.62     & \xmark & \xmark \\
DeepSeek-V3  & \underline{46.19} & 41.83 & 25.57 & 26.68 & 41.29     & \underline{44.12} & 41.38 & 25.25 & 26.29 & 40.90     & \underline{35.15} & 30.06 & \underline{12.75} & 22.95 & 28.27     & \xmark & \cmark \\
\cmidrule(lr){1-18}
Qwen-2.5 32B  & 43.56 & \underline{47.87} & 23.27 & 20.43 & \textbf{46.23} & 42.01 & \underline{46.61} & 22.48 & 19.50 & \textbf{44.63}  & 34.97 & \textbf{31.85} & 11.84 & 20.61 & \textbf{29.08} & \xmark & \cmark \\
DeepSeek-R1 32B & 37.98 & 43.29 & 12.97 & 8.87 & 42.57     & 37.14 & 42.27 & 12.71 & 8.43 & 41.45     & 32.40 & 29.92 & 10.04 & 16.31 & 28.11     & \xmark & \cmark \\
Qwen 2.5-7B   & 30.21 & 36.28 & 6.17 & 6.43 & 35.22     & 29.75 & 36.41 & 5.67 & 5.42 & 35.07     & 31.99 & 28.56 & 9.86 & 16.41 & 27.81     & \xmark & \cmark \\
DeepSeek-R1 7B  & 14.77 & 20.94 & 0.79 & 0.37 & 12.53     & 15.31 & 24.17 & 0.86 & 0.46 & 16.16     & 24.58 & 23.32 & 8.67 & 5.67 & 18.52     & \xmark & \cmark \\
NLLB-200 3.3B  & 25.05 & 25.27 & 15.86 & 20.83 & 25.30     & 24.21 & 23.64 & 15.19 & 20.13 & 22.97     & 31.45 & 25.73 & 11.49 & 18.68 & 25.51     & \cmark & \cmark \\
DeepSeek-R1 1.5B & 0.07 & 0.08 & 0.05 & 1.06 & 0.05      & 0.03 & 0.06 & 0.01 & 0.11 & 0.02      & 0.77 & 0.90 & 1.80 & 3.80 & 0.55      & \xmark & \cmark \\
M2M-100 1.2B  & 2.13 & 9.53 & 0.05 & 0.00 & 3.33      & 1.32 & 9.47 & 0.01 & 0.00 & 3.19      & 9.88 & 21.65 & 4.69 & 0.00 & 13.01     & \cmark & \cmark \\
\textbf{MERIT-3B (Ours)} & 29.71 & 34.73 & 5.15 & 4.56 & 31.20      & 27.20 & 33.16 & 4.28 & 3.54 & 29.25      & 31.91 & 26.63 & 8.40 & 13.68 & 21.18      & \cmark & \cmark \\
\midrule
\multicolumn{1}{c|}{\textbf{Semantic-Friendly}} 
& \multicolumn{5}{c|}{\textbf{chrF}}
& \multicolumn{5}{c|}{\textbf{METEOR}}
& \multicolumn{5}{c}{\textbf{BERTScore}} 
& \multicolumn{1}{c}{\textbf{}}
& \multicolumn{1}{c}{\textbf{}}\\
\cmidrule(r){1-1} \cmidrule(lr){2-16} \cmidrule(lr){17-18}
\multicolumn{1}{c|}{Model} 
& fil & id & lo & my & vi
& fil & id & lo & my & vi
& fil & id & lo & my & vi
& FT 
& OS \\
\midrule
GPT-4o     & 39.36 & 41.13 & 24.20 & 26.76 & 39.62     & 67.80 & 70.34 & 53.66 & 56.59 & \underline{69.44} & 68.59 & 70.84 & 56.27 & 57.04 & \underline{70.41}     & \xmark & \xmark \\
Claude-3.5 Sonnet & 38.71 & 41.30 & \underline{28.38} & \underline{28.79} & \underline{39.65} & 67.52 & 70.29 & \underline{58.53} & \underline{58.88} & 69.23     & 68.28 & 71.55 & \underline{60.72} & \underline{57.68} & 70.40     & \xmark & \xmark \\
Gemini-2.5 Flash& \textbf{42.68} & \textbf{41.93} & \textbf{30.61} & \textbf{32.09} & 34.66 & \textbf{70.14} & \textbf{70.87} & \textbf{60.21} & \textbf{61.85} & 60.31 & \textbf{71.67} & \textbf{72.77} & \textbf{63.93} & \textbf{62.39} & 60.88     & \xmark & \xmark \\
DeepSeek-V3  & \underline{39.80} & 36.95 & 22.37 & 24.43 & 36.86  & \underline{68.04} & 67.10 & 51.41 & 53.66 & 67.43     & \underline{70.02} & 68.81 & 54.82 & 54.21 & 68.87     & \xmark & \cmark \\
\cmidrule(lr){1-18}
Qwen-2.5 32B  & 37.77 & \underline{41.35} & 20.36 & 18.13 & \textbf{39.80} & 66.61 & \underline{70.57} & 46.72 & 43.29 & \textbf{69.76} & 67.73 & \underline{71.73} & 50.50 & 45.09 & \textbf{71.13} & \xmark & \cmark \\
DeepSeek-R1 32B & 33.62 & 37.62 & 12.77 & 9.81 & 36.97     & 61.60 & 66.75 & 34.07 & 27.37 & 66.95     & 63.32 & 68.54 & 36.09 & 27.36 & 68.65     & \xmark & \cmark \\
Qwen-2.5 7B   & 27.88 & 33.68 & 7.39 & 7.95 & 33.28     & 54.51 & 62.56 & 20.84 & 23.25 & 63.00     & 54.95 & 62.82 & 22.24 & 23.35 & 63.28     & \xmark & \cmark \\
DeepSeek-R1 7B  & 15.43 & 21.60 & 2.20 & 1.35 & 14.87     & 36.02 & 49.35 & 8.45 & 6.46 & 37.61     & 34.22 & 49.07 & 1.46 & 4.10 & 36.54     & \xmark & \cmark \\
NLLB-200 3.3B  & 22.48 & 23.57 & 14.92 & 18.69 & 22.43     & 46.10 & 45.28 & 36.27 & 42.28 & 45.73     & 48.61 & 54.34 & 44.80 & 48.93 & 52.59     & \cmark & \cmark \\
DeepSeek-R1 1.5B & 0.33 & 0.38 & 0.36 & 1.87 & 0.24      & 1.99 & 2.34 & 2.17 & 3.80 & 1.43      & 27.97& 26.07& 20.48& 19.68& 29.48     & \xmark & \cmark \\
M2M-100 1.2B  & 5.16 & 18.21 & 0.26 & 0.01 & 6.49      & 4.65 & 14.00 & 0.57 & 0.11 & 6.10      & 16.71& 1.68 & 10.04& 16.34& 9.51     & \cmark & \cmark \\
\textbf{MERIT 3B (Ours)} & 25.52 & 30.22 & 5.81 & 5.53 & 26.98      & 49.88 & 56.46 & 16.55 & 16.93 & 50.50      & 46.70 & 45.58 & 11.45 & 16.12 & 32.87      & \cmark & \cmark \\
\bottomrule
\end{tabular}
}
\vspace{-5pt}
\caption{Evaluation on five Southeast Asian languages. Strict-Overlap metrics include BLEU-4, sacreBLEU, and ROUGE-L. Semantic-Friendly metrics include chrF, METEOR, and BERTScore. For each metric column: \textbf{Bold} values indicate the highest score, and \underline{Underlined} values indicate the second highest score across all models. FT: Fine-tuned; OS: Open Source.}
\label{tab:two-block-metric}
\vspace{-2em}
\end{table*}
\autoref{tab:two-block-metric} presents our evaluation results across five LRL{\textrightarrow}Chinese directions, categorizing metrics into Strict-Overlap~\citep{papineni2002bleu, post2018call, lin2004rouge} and Semantic-Friendly~\citep{popovic2015chrf, banerjee2005meteor, zhang2019bertscore}. Among leading closed-source models, Gemini-2.5-Flash consistently achieves top scores in BLEU-4 and chrF across multiple languages, such as Filipino (BLEU-4: 49.26, chrF: 42.68) and Indonesian (BLEU-4: 48.73, chrF: 41.93). Claude-3.5 Sonnet excels in ROUGE-L for Lao (14.00) and Burmese (24.44). 

The proposed MERIT framework demonstrates notable strengths. Specifically, MERIT-3B significantly outperforms the similarly sized open-source NLLB-200-3.3B model across several metrics for Filipino, Indonesian, and Vietnamese (see \autoref{fig:experiments} and \ref{tab:significance-tests}). For instance, on Filipino, MERIT-3B achieves 29.71 BLEU-4 and 49.88 METEOR compared to NLLB-200's 25.05 and 46.10, respectively. Furthermore, MERIT-3B shows substantial gains over smaller open-source baselines on particularly challenging low-resource language pairs. Notably, our MERIT-3B model demonstrates substantial advantages over the DeepSeek-R1 7B. MERIT-3B consistently outperforms DeepSeek-R1 7B on lexical similarity metrics such as BLEU-4 and chrF across all five evaluated languages. Furthermore, when benchmarked against Qwen-2.5 7B and MERIT-3B, with only approximately 42.9\% of its parameters, MERIT-3B achieves highly competitive translation quality. For instance, on Filipino, MERIT-3B reaches 99.7\% of Qwen-2.5 7B's ROUGE-L score (31.91 vs. 31.99) and over 98.3\% of its BLEU-4 score (29.71 vs. 30.21). Similar competitiveness is observed for Indonesian, where MERIT-3B attains approximately 95.7\% of Qwen-2.5 7B's BLEU-4 score (34.73 vs. 36.28) and 93.2\% of its ROUGE-L score (26.63 vs. 28.56). For distinct scripts such as Lao and Burmese, models $\le$1.5B fail catastrophically: instruction-following deficits lead to hallucinations or source copying, yielding negligible utility.

These results underscore the efficacy of our reward-informed filtering and specialized fine-tuning approach, particularly in improving performance for low-resource languages and achieving competitive results within the open-source landscape, given model scale.

\subsection{Module Comparison}

\begin{table*}[t]
\centering
\scriptsize
\setlength{\tabcolsep}{4pt}
\vspace{-5pt}
\resizebox{\textwidth}{!}{
\begin{tabular}{c|ccccc|ccccc|c}
\toprule
\multirow{3}{*}{\textbf{Model}} 
& \multicolumn{5}{c|}{\textbf{BLEU-4}} 
& \multicolumn{5}{c|}{\textbf{chrF}} 
& \multirow{3}{*}{\makecell{\textbf{Overall}\\(BLEU-chrF)}} \\
\cmidrule(lr){2-11} 
& fil & id & lo & my & vi & fil & id & lo & my & vi & \\
\midrule
\underline{\textbf{Qwen2.5-0.5b}} & 0.03 & 0.03 & 0.02 & 0.01 & 0.01 & 0.16 & 0.12 & 0.40 & 0.25 & 0.06 & 0.11 \\
+ SFT-LTP & 1.86$_{\textcolor{aclgreen}{\uparrow1.83}}$ & 4.02$_{\textcolor{aclgreen}{\uparrow4.00}}$ & 0.25$_{\textcolor{aclgreen}{\uparrow0.22}}$ & 0.15$_{\textcolor{aclgreen}{\uparrow0.14}}$ & 3.12$_{\textcolor{aclgreen}{\uparrow3.11}}$ & 4.85$_{\textcolor{aclgreen}{\uparrow4.69}}$ & 10.38$_{\textcolor{aclgreen}{\uparrow10.26}}$ & 1.41$_{\textcolor{aclgreen}{\uparrow1.00}}$ & 1.07$_{\textcolor{aclgreen}{\uparrow0.82}}$ & 8.55$_{\textcolor{aclgreen}{\uparrow8.50}}$ & 3.57$_{\textcolor{aclgreen}{\uparrow3.46}}$ \\
+ GRPO-SAR & 2.31$_{\textcolor{aclgreen}{\uparrow2.28}}$ & 4.32$_{\textcolor{aclgreen}{\uparrow4.29}}$ & 0.26$_{\textcolor{aclgreen}{\uparrow0.24}}$ & 0.16$_{\textcolor{aclgreen}{\uparrow0.16}}$ & 6.29$_{\textcolor{aclgreen}{\uparrow6.27}}$ & 5.00$_{\textcolor{aclgreen}{\uparrow4.84}}$ & 10.64$_{\textcolor{aclgreen}{\uparrow10.52}}$ & 1.38$_{\textcolor{aclgreen}{\uparrow0.98}}$ & 1.22$_{\textcolor{aclgreen}{\uparrow0.96}}$ & 12.36$_{\textcolor{aclgreen}{\uparrow12.30}}$ & 4.39$_{\textcolor{aclgreen}{\uparrow4.28}}$ \\
+ LLME & 0.32$_{\textcolor{aclgreen}{\uparrow0.29}}$ & 0.45$_{\textcolor{aclgreen}{\uparrow0.42}}$ & 0.08$_{\textcolor{aclgreen}{\uparrow0.06}}$ & 0.07$_{\textcolor{aclgreen}{\uparrow0.06}}$ & 0.79$_{\textcolor{aclgreen}{\uparrow0.78}}$ & 1.20$_{\textcolor{aclgreen}{\uparrow1.04}}$ & 1.66$_{\textcolor{aclgreen}{\uparrow1.54}}$ & 0.45$_{\textcolor{aclgreen}{\uparrow0.05}}$ & 1.25$_{\textcolor{aclgreen}{\uparrow1.00}}$ & 2.82$_{\textcolor{aclgreen}{\uparrow2.76}}$ & 0.91$_{\textcolor{aclgreen}{\uparrow0.80}}$ \\
\makecell[c]{\textbf{Avg.}} & 1.13 & 2.21 & 0.15 & 0.10 & 2.55 & 2.80 & 5.70 & 0.91 & 0.95 & 5.95 & 2.25 \\
\midrule
\underline{\textbf{Qwen2.5-3b}} & 5.80 & 14.25 & 1.11 & 1.83 & 17.06 & 8.83 & 17.71 & 2.05 & 3.03 & 19.35 & 9.10 \\
+ SFT-LTP & 23.01$_{\textcolor{aclgreen}{\uparrow17.21}}$ & 26.00$_{\textcolor{aclgreen}{\uparrow11.75}}$ & 1.17$_{\textcolor{aclgreen}{\uparrow0.05}}$ & 2.53$_{\textcolor{aclgreen}{\uparrow0.69}}$ & 27.94$_{\textcolor{aclgreen}{\uparrow10.87}}$ & 20.14$_{\textcolor{aclgreen}{\uparrow11.31}}$ & 24.25$_{\textcolor{aclgreen}{\uparrow6.54}}$ & 2.22$_{\textcolor{aclgreen}{\uparrow0.17}}$ & 3.53$_{\textcolor{aclgreen}{\uparrow0.50}}$ & 24.55$_{\textcolor{aclgreen}{\uparrow5.20}}$ & 15.53$_{\textcolor{aclgreen}{\uparrow6.43}}$ \\
+ GRPO-SAR & 25.58$_{\textcolor{aclgreen}{\uparrow19.78}}$ & 29.11$_{\textcolor{aclgreen}{\uparrow14.86}}$ & 4.39$_{\textcolor{aclgreen}{\uparrow3.28}}$ & 2.77$_{\textcolor{aclgreen}{\uparrow0.93}}$ & 32.54$_{\textcolor{aclgreen}{\uparrow15.48}}$ & 23.62$_{\textcolor{aclgreen}{\uparrow14.78}}$ & 27.83$_{\textcolor{aclgreen}{\uparrow10.12}}$ & 5.17$_{\textcolor{aclgreen}{\uparrow3.12}}$ & 3.45$_{\textcolor{aclgreen}{\uparrow0.42}}$ & 27.88$_{\textcolor{aclgreen}{\uparrow8.53}}$ & 18.23$_{\textcolor{aclgreen}{\uparrow9.13}}$ \\
+ LLME & 29.71$_{\textcolor{aclgreen}{\uparrow23.91}}$ & 34.73$_{\textcolor{aclgreen}{\uparrow20.48}}$ & 5.15$_{\textcolor{aclgreen}{\uparrow4.04}}$ & 4.56$_{\textcolor{aclgreen}{\uparrow2.73}}$ & 31.20$_{\textcolor{aclgreen}{\uparrow14.14}}$ & 25.52$_{\textcolor{aclgreen}{\uparrow16.69}}$ & 30.22$_{\textcolor{aclgreen}{\uparrow12.51}}$ & 5.81$_{\textcolor{aclgreen}{\uparrow3.76}}$ & 5.53$_{\textcolor{aclgreen}{\uparrow2.50}}$ & 26.98$_{\textcolor{aclgreen}{\uparrow7.63}}$ & 19.94$_{\textcolor{aclgreen}{\uparrow10.84}}$ \\
\makecell[c]{\textbf{Avg.}} & 21.03 & 26.02 & 2.96 & 2.92 & 27.19 & 19.53 & 25.00 & 3.81 & 3.89 & 24.69 & 15.70 \\
\bottomrule
\end{tabular}}
\caption{Evaluation results of Qwen-2.5-0.5B and 3B on five Southeast Asian languages. All values are rounded to two decimal places. Improvements over the baseline (\underline{underlined rows}) are shown with arrows.}
\label{tab:qwen25-results}
\vspace{-2em}
\end{table*}

To investigate the contribution of each component, we conduct an ablation study on Qwen-2.5-0.5B and Qwen-2.5-3B across four distinct setups: zero-shot (serving as our baseline), Supervised Fine-Tuning with Language-Token Prefixing (SFT-LTP), SFT-LTP followed by reward-enhanced tuning using Group Relative Policy Optimization with Semantic Alignment Reward (GRPO-SAR), and finally our full SFT-LTP + GRPO-SAR with an additional LLMs-Enhanced (LLME) stage.

As detailed in Table~\ref{tab:qwen25-results}, initial SFT-LTP yields substantial improvements in both BLEU-4 and chrF scores over the zero-shot baselines across all languages for both model sizes. For instance, Qwen-2.5 3B sees its overall BLEU-chrF score increase from 9.10 to 15.53 after SFT-LTP.
Introducing GRPO-SAR provides further consistent gains. Notably, for the Qwen-2.5 3B model, GRPO-SAR significantly boosts performance on low-resource pairs like Lao{\textrightarrow}Chinese, improving BLEU-4 from 1.17 (SFT-LTP) to 4.39 and chrF from 2.22 (SFT-LTP) to 5.17. Even with its limited capacity, the Qwen-2.5 0.5B model benefits remarkably from GRPO-SAR, achieving an overall BLEU-chrF score of 4.39, which is nearly a 40-fold increase (a 3890\% relative improvement) over its zero-shot baseline score of 0.11. This underscores the efficacy of reward modeling, consistent with findings in instruction tuning \citep{ouyang2022training, wu2023finegrained}.

Our proposed LLMs-Enhanced (LLME) stage demonstrates further advancements, particularly for the larger Qwen-2.5 3B model. With LLME, the Qwen-2.5 3B model achieves the highest overall BLEU-chrF score of 19.94, representing a 10.84 absolute point improvement (a 119\% relative increase) over its zero-shot baseline. This highlights the synergistic benefits of our full pipeline. While the LLME stage yields more modest gains for the Qwen-2.5 0.5B model in the current setup (overall BLEU-chrF of 0.91), the substantial cumulative improvements from SFT-LTP and GRPO-SAR on this smaller model, and the peak performance achieved by the 3B model with LLME, collectively validate the effectiveness and scalability of our modular tuning strategy in significantly enhancing translation quality.

\subsection{Effect of Data Distillation on Performance}

We assess the impact of our quality filtering approach by comparing full-scale Supervised Fine-Tuning with Language-Token Prefixing (SFT-LTP) against subsequent reward-informed filtering and tuning via GRPO-SAR, using our MERIT-3B model. \autoref{tab:training-size-score} details the number of retained training instances per language and the corresponding overall BLEU-chrF scores for these configurations.

The SFT-LTP stage utilizes the full set of 40,000 training instances. In contrast, the GRPO-SAR stage strategically curates this data, drastically reducing the volume to only 9,126 instances. This results in an average data reduction of 77.2\%, with the most significant reduction observed for Vietnamese, where the training data was reduced by 87.8\% (from 8,000 to 976 instances). Remarkably, despite this substantial data pruning, the overall BLEU-chrF score not only signifies the efficient retention of highly informative samples but actually improves from 15.53 (achieved with SFT-LTP on 40,000 instances) to 18.23 with GRPO-SAR on the reduced dataset. This represents a relative performance increase of approximately 17.4\%.

These findings underscore the efficacy of our reward-based filtering as a data-efficient strategy that simultaneously reduces training data requirements and enhances model performance. This offers a compelling alternative to training on larger, potentially noisier, unfiltered datasets. The benefits of leveraging reward signals for targeted data curation align with effective strategies observed in other generative AI tasks, such as summarization and dialogue tuning \citep{ouyang2022training}.

\section{Discussion}

Future work should incorporate robust script normalization or transliteration to mitigate encoding inconsistencies in Lao and Burmese. The current GRPO-SAR reward may overemphasize adequacy; multi-objective rewards that balance adequacy and fluency are needed. Zero-shot baselines of large closed-source LLMs may be underestimated; combining our data distillation with advanced prompting or in-context learning on larger models remains a promising direction. The full discussion is visible in Section \ref{sec:disscusion}.

\section{Conclusion}
This work bridges the critical gap in Chinese-centric low-resource translation by introducing MERIT, a comprehensive framework comprising the CALT benchmark and a data-efficient training paradigm. By synergizing Language-specific Token Prefixing (LTP), Supervised Fine-Tuning (SFT), and our novel Group Relative Policy Optimization (GRPO) guided by the Semantic Alignment Reward (SAR), we demonstrate that model performance can be significantly decoupled from sheer parameter scale. Notably, our MERIT-3B model surpasses much larger baselines, such as NLLB-200 3.3B and M2M-100, while utilizing only 22.8\% of the original training data. These findings underscore that in low-resource regimes, expert-aligned data distillation and reward-guided optimization offer a more sustainable and effective path than brute-force scaling, providing a reproducible blueprint for narrowing the multilingual digital divide.


\newpage

\section*{Limitations}

Despite the encouraging results achieved by our proposed framework and the MERIT-3B model, this work has several limitations that warrant discussion and offer avenues for future improvement.

\textbf{Limited Linguistic Coverage:} Despite constructing direct LRL-Chinese sentence pairs, the current benchmark is restricted to only five Southeast Asian languages, leaving important low-resource languages such as Tibetan, Uyghur, and Kazakh unaddressed due to the lack of high-quality parallel corpora.

\textbf{Residual English-Centric Bias in the Test Suite:} Although the ALT-based test suite has been realigned for LRL-Chinese evaluation using shared alt\_id indexing, it remains inherently constrained by its original English-centric design, potentially retaining subtle domain-specific or stylistic artifacts that affect translation assessment.

\textbf{Insufficient Scale of Human Validation:} While the QE agent and statistical-semantic SAR function rely on automatic filtering, the volume of human validation, particularly the expert-annotated data employed in reward model training, remains substantially limited. This constraint may compromise the robustness and alignment of the SAR model with diverse human preferences.

\textbf{Fixed Decoding and Prompting Strategies:} Although multiple LLMs were evaluated under zero-shot, SFT, and GRPO settings, decoding hyperparameters (e.g., beam size, temperature) and prompt formats were uniformly fixed for fair comparison, potentially masking model-specific performance differences that could be uncovered through more extensive hyperparameter and prompt engineering.

\textbf{Restricted Model Scale and Missing Efficiency Analysis:} Due to computational constraints, all experiments including the MERIT-3B model and the proposed SFT-LTP and GRPO-SAR frameworks were limited to models of up to 3B parameters; the approach has not yet been validated on larger (over 7B) or state-of-the-art billion-scale models, and a comprehensive efficiency analysis of training time, inference latency, and overall computational cost remains absent, leaving scalability and practical deployability unassessed.

\section*{Ethics Statements}

This work presents a Chinese-centric multilingual translation benchmark targeting five Southeast Asian low-resource languages (LRLs), constructed from publicly available corpora and evaluated under reproducible protocols. We aim to support responsible research in multilingual NLP by releasing rigorous evaluation resources while proactively addressing ethical concerns related to data provenance, model fairness, environmental impact, and potential misuse.

\paragraph{Data Privacy and Consent} 
All data is derived from the publicly available ASEAN Languages Treebank (ALT), which includes multilingual translations of government and news texts. While the dataset is openly licensed, the original collection did not explicitly document consent procedures or procedures for removing personally identifiable information (PII). To mitigate this, we apply a multi-stage filtering process to exclude named entities, explicit language, and potentially sensitive content. Nonetheless, due to the limitations of automated and manual filtering, some residual risk may remain. We follow the data statements framework \citep{bender2018data} and document licensing, provenance, and usage constraints in the appendix.

\paragraph{Bias and Fairness} 
Despite the use of a three-stage filtering pipeline and expert-rated supervision, the training data may still encode latent cultural, linguistic, or regional bias, particularly due to its English-pivoted design and limited coverage of dialectal variations or non-standard orthographies. Annotators are bilingual graduate students, and while they are experienced, demographic diversity is limited. Future work will prioritize the inclusion of more diverse annotators and typologically broader sources to mitigate such representational imbalances. Our work aligns with global AI ethics principles of fairness, transparency, and non-maleficence \citep{gebru2018datasheets}.

\paragraph{Environmental Impact} 
Model training and inference were conducted on two NVIDIA RTX 3090 GPUs (24 GB). We log training FLOPs and wall-clock runtime for both the SFT and GRPO stages. While the GRPO procedure improves data efficiency through reward-based filtering, it introduces additional computational cost. We estimate that the total training corresponds to a typical single-node compute workload and plan to explore more lightweight reward models or compute-efficient alternatives to reduce carbon impact in future iterations \citep{emnlp2023ethicsfaq,lu2026attention,li2026ecothinkgreenadaptiveinference}.

\paragraph{Intended Use and Misuse Risks} 
The benchmark is designed to support objective evaluation and supervised training for LRL{\textrightarrow}Chinese translation tasks. It is intended for academic research and language technology development, particularly in regions underrepresented in NLP. However, misuse is possible, such as generating misinformation or content targeting marginalized communities. We explicitly discourage such applications and recommend that any downstream use include fairness auditing, risk controls, and human oversight \citep{mitchell2018modelcards}.


\bibliography{intro}

\clearpage
\appendix
\section{Appendix}

\subsection{Further Discussion}
\label{sec:disscusion}
Our work, culminating in the MERIT framework, demonstrates the significant potential of combining data filtering techniques, such as the Semantic Alignment Reward (SAR) driven GRPO, with efficient fine-tuning strategies like Language-Token Prefixing (LTP) for multilingual translation, especially into low-resource languages (LRLs). The proposed BLEU-chrF composite metric has also provided a balanced view of lexical and semantic performance. While MERIT-3B exhibits strong performance relative to its scale and against comparable open-source models, several limitations persist and pave the way for future exploration.

First, script-related challenges, particularly for Lao and Burmese, can introduce encoding inconsistencies. These not only affect the performance of QE agents used in SAR but also potentially skew standard evaluation metrics. Future iterations could incorporate more robust character normalization or transliteration techniques at the data preprocessing stage, or develop QE models less sensitive to such variations.

Second, the current reward model underlying GRPO-SAR, while effective, may inadvertently prioritize adequacy (accuracy of content, as captured by our specific SAR function focusing on numerical or key information matching) sometimes at the expense of optimal fluency. This can occasionally lead to subtle grammatical artifacts in some translations. Future work could investigate multi-objective reward functions that explicitly balance adequacy, fluency, and even other aspects like style or register, potentially drawing on more diverse human feedback signals beyond simple ratings.

As noted by \citet{zhang2022opt}, many LLMs, including some baselines we compared against, are often evaluated in zero-shot or few-shot settings for translation. This might not fully reveal their capabilities, which could be significantly enhanced with more sophisticated prompting strategies or in-context learning techniques specifically tailored for translation. Exploring how our data filtering and fine-tuning methods can synergize with advanced prompting for even larger LLMs is a promising direction.
\subsection{Supplementary Experiments}
All experiments were conducted on a local workstation equipped with two NVIDIA RTX 3090 GPUs (24 GB). Under a 2×2 parallel configuration, the per-GPU batch size was set to 8 with a gradient accumulation step of 2, resulting in an effective total batch size of 32. The maximum input sequence length was set to 1024 tokens, and the initial learning rate was configured as \texttt{2e-4}. The system environment included Ubuntu 20.04, CUDA 12.1, and Python 3.10, with PyTorch 2.1 and Transformers v4.49 as the core libraries. All training was performed using standard mixed-precision (fp16) computation via custom training scripts. Due to hardware limitations, the batch size was carefully adjusted to fit within the available GPU memory, and no experiments were conducted using larger-parameter models. To ensure reproducibility, all random seeds were fixed, and detailed runtime logs were maintained for each experiment.

\begin{table*}
\centering
\renewcommand{\arraystretch}{1.2}
\setlength{\tabcolsep}{6pt}
\small
\begin{tabular}{c|ccccc|c}
\toprule
\multirow{3}{*}{\textbf{Method}} & \multicolumn{5}{c|}{\textbf{\# Training Size}} & \makecell{\textbf{Overall}} \\
& \textbf{fil} & \textbf{id} & \textbf{lo} & \textbf{my} & \textbf{vi} & \textbf{(Size / BLEU-chrF)}\\
\midrule
{\textbf{MERIT-3B}} & & & & & & \\ 

+ SFT-LTP & \textbf{8,000} & \textbf{8,000} & \textbf{8,000} & \textbf{8,000} & \textbf{8,000} & \textbf{40,000 / 15.53} \\ 

+ GRPO-SAR & 
1,851$_{\textcolor{aclred}{\downarrow76.9\%}}$ & 
1,779$_{\textcolor{aclred}{\downarrow77.7\%}}$ & 
2,058$_{\textcolor{aclred}{\downarrow74.2\%}}$ & 
2,462$_{\textcolor{aclred}{\downarrow69.2\%}}$ & 
976$_{\textcolor{aclred}{\downarrow87.8\%}}$ & 
9,126$_{\textcolor{aclred}{\downarrow77.2\%}}$ / 18.23$_{\textcolor{aclgreen}{\uparrow17.4\%}}$ \\

+ LLME & 
2,891$_{\textcolor{aclred}{\downarrow63.9\%}}$ & 
3,104$_{\textcolor{aclred}{\downarrow61.2\%}}$ & 
3,300$_{\textcolor{aclred}{\downarrow58.8\%}}$ & 
3,764$_{\textcolor{aclred}{\downarrow53.0\%}}$ & 
2,193$_{\textcolor{aclred}{\downarrow72.6\%}}$ & 
15,252$_{\textcolor{aclred}{\downarrow61.9\%}}$ / 19.94$_{\textcolor{aclgreen}{\uparrow28.4\%}}$ \\

\bottomrule
\end{tabular}
\vspace{-0.5em}
\caption{Training size comparison across five low-resource languages for MERIT-3B. Overall column shows total training data (with percentage reduction relative to initial 40,000) and BLEU-chrF score (with percentage improvement relative to the SFT-LTP stage).}
\label{tab:training-size-score}
\end{table*}
\begin{figure*}
 \includegraphics[width=0.49\linewidth]{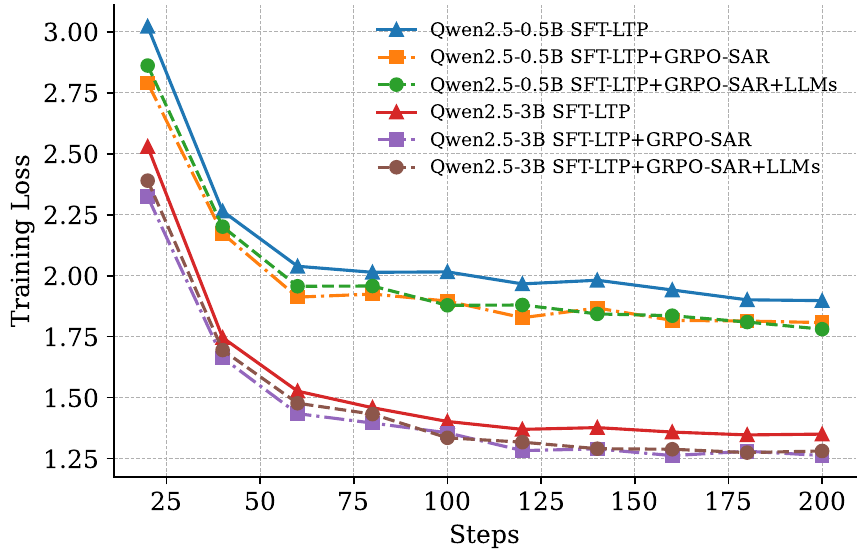} \hfill
 \includegraphics[width=0.49\linewidth]{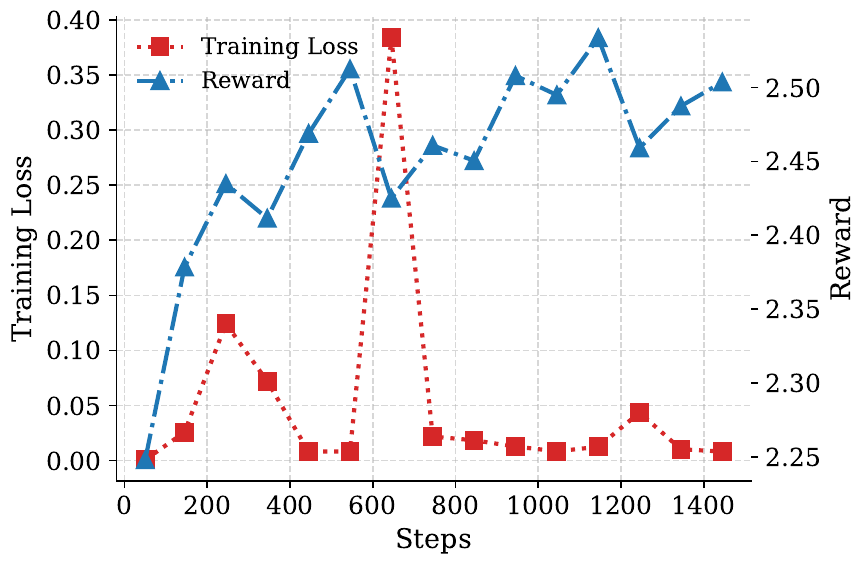}
 \vspace{-1em} 
 \caption {Training loss and reward evolution across SFT and GRPO strategies.}
\end{figure*}

\subsection{Reward Function}
In this study, we introduce the Semantic Alignment Reward (SAR) function as a key component of the reward mechanism for aligning the Quality Estimation (QE) agent's predictions with human expert judgments. Specifically, based on the absolute deviation $d = |s_i - a_i|$ between the predicted score $s_i$ and the ground-truth expert score $a_i$, we design a step-wise reward strategy which operates as follows:

\begin{itemize}
\item A reward of \textbf{2.0} is assigned if the model’s predicted score exactly matches the human expert score ($d=0$).

\item A reward of \textbf{1.0} is assigned if the predicted score deviates from the expert score by an acceptable margin (specifically, $1 \le d \le 10$).

\item A reward of \textbf{0.0} is assigned if the deviation exceeds the tolerance threshold ($d > 10$) or if the score is invalid.
\end{itemize}

This step-wise formulation distinguishes SAR from binary rewards typical in exact matching tasks, driven by two key rationales:

\textbf{(1) Accommodating Evaluation Subjectivity.}
Unlike mathematical reasoning with unique solutions, QE involves inherent subjectivity where human annotators may assign varying scalar scores to identical quality levels. Consequently, enforcing exact matches is overly rigid. Our tolerance mechanism acknowledges this by treating proximal scores (within a 10-point window) as valid, thereby preventing penalization for minor variances that do not reflect genuine quality disparities.

\textbf{(2) Providing Dense Supervision Signals.}
Sparse binary rewards often hinder optimization stability. By granting partial rewards for approximate matches, SAR provides finer-grained feedback even when exact alignment is not achieved. This dense signal facilitates smoother convergence during GRPO, guiding the agent to approximate the distribution of expert preferences rather than overfitting to rigid scalar values.



\begin{table}
\small
\centering
\begin{tabular}{lcc}
\toprule
\textbf{Combination} & \textbf{Spearman $\rho$ $\uparrow$} & \textbf{$\sigma^2$ $\downarrow$} \\
\midrule
$($BLEU $+$ chrF$)/2$ (equal) & \textbf{0.986} & \textbf{1.02} \\
$\sqrt{\text{BLEU} \times \text{chrF}}$ (geometric) & 0.985 & 1.14 \\
$0.4 \cdot \text{BLEU} + 0.6 \cdot \text{chrF}$ & 0.985 & 1.19 \\
$0.6 \cdot \text{BLEU} + 0.4 \cdot \text{chrF}$ & 0.984 & 1.23 \\
\bottomrule
\end{tabular}
\parbox{0.9\linewidth}{\small \textit{Note:} Metrics were computed with sacreBLEU 2.4.2.}
\caption{Empirical comparison of different combination methods for BLEU and chrF, evaluated on WMT22 test sets (5 languages $\times$ 16 systems). The results show that, while all variants exhibit very similar rank correlations with human judgments, the simple arithmetic mean yields the lowest system-level variance.}
\label{tab:metric-comparison}
\vspace{-1em}
\end{table}

\begin{table}
\small
\centering
\begin{tabular}{lcc}
\toprule
\textbf{Metric} & \textbf{Mean $\Delta$ (v.s NLLB-200)} & \textbf{p-value} \\
\midrule
BLEU-4   & +9.46 & \textbf{0.003} \\
chrF2    & +8.36 & \textbf{0.006} \\
ROUGE-L   & +1.18 & \textbf{0.041} \\
METEOR   & +2.65 & \textbf{0.038} \\
COMET-22  & +2.80 & \textbf{$<$0.001} \\
\bottomrule
\end{tabular}
\parbox{0.9\linewidth}{\small \textit{Note:} Mean $\Delta$ represents the score difference averaged across all test sets. Some metrics are presented as percentages. p-values in \textbf{bold} indicate a statistically significant improvement ($p < 0.05$).}
\caption{Statistical Significance of Improvements over Baseline. We conduct pairwise bootstrap resampling (1,000 iterations) to compare our final model (MERIT-3B) against the NLLB-200 baseline. All improvements are statistically significant.}
\label{tab:significance-tests}
\vspace{-2em}
\end{table}

\subsection{Recruitment And Payment}
To ensure the accuracy and objectivity of human evaluation, we recruited ten annotators with academic backgrounds in the target Southeast Asian languages. All annotators were either language instructors or graduate students from relevant universities. For each target language, three annotators were assigned, and a cross-review protocol was adopted to enhance annotation quality and consistency. All participants had formal training in translation or linguistics and possessed strong language comprehension and evaluative capabilities. Annotators were compensated at a rate of 1 RMB per evaluated sample. Before the evaluation began, all participants received detailed instructions and training on annotation guidelines. Participation was voluntary, and compensation was provided proportionally based on the amount of completed work. Since the dataset contains no personally identifiable information (PII) and the task involves only linguistic quality assessment, the annotation process entails no ethical risks and does not require institutional ethics approval.
\begin{algorithm}[b!]
\caption{Elite Parallel Data Sampler}
\label{alg:epds}
\KwData{$\mathcal{D}_{raw}$, Target Size $K$, LLM $\mathcal{M}$}
\KwResult{$\mathcal{D}_{clean}$}

$\mathcal{D}_{valid} \gets \emptyset$\;
\ForEach{$(x, y) \in \mathcal{D}_{raw}$}{
  \If{$\mathbb{I}_{filter}(x, y) = 1$}{
    $\mathbf{f}_{stat} \gets \text{ExtractFeatures}(x, y)$\;
    $S_{base} \gets \mathbf{w}^T \cdot \Phi(\mathbf{f}_{stat})$\;
    $\mathcal{D}_{valid} \gets \mathcal{D}_{valid} \cup \{(x, y, S_{base})\}$\;
  }
}

\ForEach{batch $B \subset \mathcal{D}_{valid}$}{
  $\mathbf{S}_{PPL}, \mathbf{S}_{IFD} \gets \mathcal{M}.\text{Forward}(B)$\;
  \ForEach{$(x, y) \in B$}{
    $S_{final}^{(x,y)} \gets S_{base} + S_{PPL}^{(x,y)} + S_{IFD}^{(x,y)}$\;
  }
}

$\mathcal{D}_{sorted} \gets \text{Sort}(\mathcal{D}_{valid}, \text{key}=S_{final})$\;
$\mathcal{D}_{clean} \gets \{d \in \mathcal{D}_{sorted} \mid \text{rank}(d) \le K\}$\;

\Return $\mathcal{D}_{clean}$\;
\end{algorithm}

\subsection{Data Filtering Methodology}
\label{sec:filtering_method}

To ensure the quality of the Chinese-centric low-resource corpora, we implement a hybrid filtering pipeline combining statistical heuristics and Large Language Model (LLM) based metrics. Let $\mathcal{D}_{raw} = \{(x_i, y_i)\}_{i=1}^N$ denote the initial noisy parallel corpus.

\subsubsection{Statistical Feature Extraction}
For each pair $(x_i, y_i)$, we extract a feature vector $\mathbf{f}_i \in \mathbb{R}^5$ measuring surface alignment:
\begin{enumerate}
  \item \textbf{Length Ratio ($R_{len}$):} defined as $\min(|y_i|/|x_i|, |x_i|/|y_i|)$ to penalize extreme length mismatches.
  \item \textbf{Token Ratio ($R_{tok}$):} Calculated on whitespace-separated tokens.
  \item \textbf{Punctuation Divergence ($D_{punct}$):} Absolute difference in punctuation ratios, adjusted by language-specific regex patterns.
  \item \textbf{Digit Divergence ($D_{digit}$):} Absolute difference in digit proportions.
  \item \textbf{Lexical Diversity Diff ($D_{uniq}$):} Difference in Type-Token Ratios (TTR).
\end{enumerate}

The base statistical score is a weighted sum with normalizing functions $\phi_k$:
\begin{equation}
  S_{base}(x_i, y_i) = \sum_{k=1}^5 w_k \cdot \phi_k(\mathbf{f}_i^{(k)})
\end{equation}

\subsubsection{LLM-based Semantic Scoring}
We employ \texttt{Qwen-2.5-0.5B} to compute semantic metrics.
\textbf{Perplexity Score ($S_{PPL}$)} verifies fluency via conditional likelihood, and \textbf{Instruction-Following Discrepancy ($S_{IFD}$)} measures context reliance:
\begin{align}
  S_{PPL} &= \frac{1}{1 + \sigma(\text{PPL}(y|x, \mathcal{I}))} \\
  S_{IFD} &= \min\left(\frac{\text{PPL}(y|\emptyset)}{\text{PPL}(y|x, \mathcal{I})}, \tau\right) \cdot \tau^{-1}
\end{align}
where $\mathcal{I}$ is the instruction prompt, $\sigma$ is a scaling factor, and $\tau$ is a normalization threshold.

\subsubsection{Composite Filtering Algorithm}
The final score combines signals with weights $(\alpha, \beta, \gamma) = (0.3, 0.3, 0.4)$:
\begin{equation}
  S_{final} = \alpha S_{base} + \beta S_{PPL} + \gamma S_{IFD}
\end{equation}

\begin{algorithm}[t!]
\caption{Data Integrity Validation}
\label{alg:div}
\KwData{$\mathcal{D}_{clean}$, Targets $\{N_{train}, N_{dev}, N_{test}\}$}
\KwResult{Splits $\{S_{train}, S_{dev}, S_{test}\}$}

$\mathcal{G} \gets \{ (k, \{s \mid d(s)=k\}) \}_{k=1}^M$\;
$\mathbf{P}[k] \gets |\mathcal{G}[k]| / |\mathcal{D}_{clean}|, \quad \forall k$\;

\ForEach{split $T \in \{train, dev, test\}$}{
  $S_T \gets \emptyset$\;
  \ForEach{domain $k \in \text{Keys}(\mathcal{G})$}{
    $q_k \gets \lfloor N_T \cdot \mathbf{P}[k] \rfloor$\;
    $B_k \gets \text{Sample}(\mathcal{G}[k], q_k)$\;
    $S_T \gets S_T \cup B_k$; \quad $\mathcal{G}[k] \gets \mathcal{G}[k] \setminus B_k$\;
  }
  
  $\Delta \gets N_T - |S_T|$\;
  \If{$\Delta > 0$}{
    $U \gets \bigcup_{k} \mathcal{G}[k]$ 
    $S_{supp} \gets \text{Resample}(U, \Delta)$\;
    $S_T \gets S_T \cup S_{supp}$\;
  }
  \ElseIf{$|S_T| > N_T$}{
    $S_T \gets S_T[1:N_T]$\;
  }
}

\lIf{$\exists T, |S_T| \neq N_T$}{\Return \textbf{Error}}
\Return $\{S_{train}, S_{dev}, S_{test}\}$\;
\end{algorithm}



\subsection{Data Integrity Validation}
\label{sec:div_method}

To ensure the reliability of our benchmark, we propose the Data Integrity Validation (DIV) algorithm. This mechanism addresses two critical challenges in low-resource corpus construction: (1) \textbf{Distributional Integrity}, ensuring that the domain distribution (e.g., news, government docs) in the training and test sets strictly mirrors the original corpus; and (2) \textbf{Size Integrity}, enforcing exact sample counts (e.g., 8,000/1,000/1,000) despite rounding errors inherent in stratified sampling.

Let $\mathcal{D}_{clean}$ be the filtered corpus. For each sentence pair $s$, we extract its domain identifier $d(s)$ from the metadata.

\subsubsection{Distribution-Preserving Allocation}
We first compute the global domain distribution vector $\mathbf{P} = \{p_1, \dots, p_M\}$, where $p_k = N_k / N_{total}$ denotes the proportion of domain $k$. For a target split $T$ (e.g., Train) with a required size $N_{target}$, the allocated quota for domain $k$ is calculated as:
\begin{equation}
  q_T^{(k)} = \lfloor N_{target} \cdot p_k \rfloor
\end{equation}
We then sample $q_T^{(k)}$ distinct sentences from domain $k$ to form the initial split.

\subsubsection{Integrity Compensation}
Since the floor operation $\lfloor \cdot \rfloor$ may cause the total count to fall short of $N_{target}$ (i.e., $\sum q_T^{(k)} < N_{target}$), or data scarcity in specific domains may prevent fulfilling the quota, DIV performs a \textit{Integrity Compensation} step. We calculate the deficit $\Delta = N_{target} - \sum |S_T|$ and perform weighted resampling from the available pool to fill $\Delta$, ensuring the final dataset size strictly equals $N_{target}$ without violating domain diversity.

\end{document}